\pdfoutput=1

\documentclass[11pt]{article}

\usepackage{acl}

\usepackage{times}
\usepackage{latexsym}
\usepackage{algorithm}
\usepackage{multirow}
\usepackage{algpseudocode}

\usepackage{graphicx,url}
\usepackage[normalem]{ulem}
\useunder{\uline}{\ul}{}

\usepackage{xcolor}
\usepackage[T1]{fontenc}

\usepackage[utf8]{inputenc}

\usepackage{microtype}

\usepackage{inconsolata}
\usepackage{booktabs}
\usepackage{multirow}
\usepackage{graphicx}
\title{From Random to Informed Data Selection: A Diversity-Based Approach to Optimize Human Annotation and Few-Shot Learning}

\author{
    Alexandre Alcoforado$^1$, Thomas Palmeira Ferraz$^2$, Lucas Hideki Okamura$^1$, \\ \textbf{Israel Campos Fama}$^1$\textbf{, Arnold Moya Lavado}$^1$ \textbf{, Bárbara Dias Bueno}$^1$\textbf{,} \\ \textbf{Bruno Veloso}$^3$\textbf{, Anna Helena Reali Costa}$^1$ \\ \hfill \\ 
     $^1$Escola Politécnica, Universidade de São Paulo, São Paulo, Brazil \\
    $^2$Télécom Paris, Institut Polytechnique de Paris, Palaiseau, France \\
    $^3$Faculty of Economics, University of Porto and INESC TEC, Porto, Portugal \\ 
    \tt \{alexandre.alcoforado, bruno.miguel.veloso, dev.arn.ml\}@gmail.com \\
    \tt thomas.palmeira@telecom-paris.fr \quad lucasokamura@alumni.usp.br \\ \tt \{israelfama, barbarabueno, anna.reali\}@usp.br
}

\begin{document}
\maketitle

\begin{abstract}
A major challenge in Natural Language Processing is obtaining annotated data for supervised learning. An option is the use of crowdsourcing platforms for data annotation. However, crowdsourcing introduces issues related to the annotator's experience, consistency, and biases. An alternative is to use zero-shot methods, which in turn have limitations compared to their few-shot or fully supervised counterparts. Recent advancements driven by large language models show potential, but struggle to adapt to specialized domains with severely limited data. The most common approaches therefore involve the human itself randomly annotating a set of datapoints to build initial datasets. But randomly sampling data to be annotated is often inefficient as it ignores the characteristics of the data and the specific needs of the model. The situation worsens when working with imbalanced datasets, as random sampling tends to heavily bias towards the majority classes, leading to excessive annotated data. To address these issues, this paper contributes an automatic and informed data selection architecture to build a small dataset for few-shot learning. Our proposal minimizes the quantity and maximizes diversity of data selected for human annotation, while improving model performance.

\end{abstract}

\section{Introduction}

In real-life scenarios, particularly in the realm of Machine Learning (ML) in Natural Language Processing (NLP), annotated data is often a scarce and challenging resource to acquire. In many cases, researchers and practitioners are faced with the daunting task of developing accurate models with extremely limited or even non-existent annotated training data. To address this challenge, the process is typically initiated by building a small annotated dataset and using it as a basis for training ML models using supervised learning methods.  
Subsequently, this process can be iterated by creating annotated datasets of increasing size through techniques commonly referred to as Active Learning (AL) \citep{ren2021survey}.

As an alternative approach to acquiring annotated data, crowdsourcing platforms like Amazon Mechanical Turk have been used in recent years. However, relying solely on human annotation services from these platforms brings its own set of challenges \citep{nowak2010reliable,karpinska2021perils}. Variability in expertise among annotators often results in inconsistent annotation criteria and, at times, conflicting annotations. Moreover, human annotators may encounter difficulties when dealing with large datasets, leading to errors and delays in data annotation processes. An additional concern lies in the potential introduction of bias through annotators' subjectivity and personal biases, which can negatively affect the performance of trained models. To mitigate these challenges, numerous research works have attempted to address these issues, either by selecting high-quality annotators in multiple-annotated-data setups or by employing diverse methods to weight each annotator's input \citep{zhang2023needle,hsueh2009data,hovy2013learning,basile2021probabilistic}. 


In low-resource settings, a common practice is to randomly sample a subset of the unlabeled data for the annotation process \citep{setfit,beijbom2014random}. This approach involves selecting a few examples at random, which are then annotated to form the initial training dataset. However, this methodology may be suboptimal since it neglects the specific characteristics of the data and the requirements of the learning model. In other words, randomly sampled data may fail to adequately represent the full spectrum of classes or concepts present within the dataset.

The advent of zero-shot methods has provided an intriguing approach to perform initial annotation without any annotated training data \citep{alcoforado2022zeroberto}. Nonetheless, historical shortcomings have often placed zero-shot methods behind their few-shot counterparts in terms of performance. Recent strides in the field of NLP, particularly the emergence of general-purpose Large Language Models (LLMs), have opened up exciting avenues in multi-task learning and zero-shot problem-solving \citep{ferraz2023distilwhisper}. These models exhibit remarkable skills across various tasks \citep{brown2020language,touvron2023llama} but still encounter difficulties when adapting to specific domains where highly specialized knowledge may be entirely absent from their training data \citep{yang2023harnessing,zhang2023reformulating}.

In the realm of few-shot text classification, the challenge of acquiring annotated data becomes increasingly daunting, particularly when confronted with imbalanced datasets \citep{ferraz2021debacer}. Common benchmark datasets used for few-shot text classification tasks often exhibit a semblance of balance or slight imbalance. However, such datasets represent rare exceptions in the real-world landscape, where data distributions are typically skewed and imbalanced, mirroring the inherent complexity of practical scenarios. The prevalence of imbalanced data poses a significant challenge, as traditional random sampling strategies become increasingly suboptimal. In scenarios where one class overwhelmingly dominates, random sampling tends to favor the majority class, resulting in data selection that inadequately represents the underrepresented and rare classes.

To address these challenges, in this paper we introduce an innovative automatic data selection architecture for few-shot learning. Our approach is designed to identify the most informative and representative data points that should be annotated by humans in low-resource, annotation-scarce scenarios. It leverages a framework that systematically orders data points based on their likelihood to (i) belong to distinct classes, thereby avoiding unnecessary redundancy in human annotation efforts, and (ii) enhance the overall performance of the learning model. Our evaluation of this approach encompasses various low-resource natural language processing datasets, demonstrating its capacity to minimize redundancy in human annotation efforts and improve model performance compared to traditional random sampling or manual data selection strategies, particularly in cases with a limited number of annotated examples.

In summary, this work presents two primary contributions:

\begin{enumerate}
\item The introduction of an automatic data selection architecture for few-shot learning that leverages active learning principles to identify the most informative and representative data points for annotation.
\item An extensive analysis of various implementations of our architecture, highlighting its effectiveness to build the first version of a dataset in the context of low-resource text classification. 
\end{enumerate}

Our results emphasize the benefits of informed data selection, which not only streamlines the annotation process but also results in a more diverse set of annotated data. Furthermore, models trained with these diverse datasets exhibit improved performance, which may benefit subsequential iterations of the dataset with Active Learning techniques. Our experiments unveil the potential of informed data selection strategies in addressing the challenges of few-shot learning in low-resource NLP scenarios.

\section{Background}

In low-resource NLP settings, where annotated data is scarce and expensive to obtain, Active Learning (AL) \citep{ren2021survey} methods show themselves as a very promising approach. AL attempts to maximize the performance gain of a model by annotating the smallest number of samples. AL algorithms select data from an unlabeled dataset and query a human annotator only on this selected data, which aims to minimize human efforts in annotation by using only the most informative data.

Uncertainty sampling \citep{zhu2010} is among the most used method to select which points to be annotated.
It employs a single classifier to pinpoint unlabeled instances where the classifier exhibits the lowest confidence.
Other approaches include query-by-committee \citep{kee2018}, where a pool of models is used to find diverse disagreements, margin sampling \citep{ducoffe2018adversarial}, and entropy sampling \citep{LiICM2011}. The first one looks for points where models disagree the most on the predicted labels; while the second selects data points with the highest entropy, indicating the lowest classification probability across all potential classes

An essential aspect of AL involves the allocation of annotation budgets. Given that human effort is dedicated to annotating data, it is crucial to maximize its utility and minimize human effort. Various strategies have emerged to address this challenge. Recent research suggests optimizing directly for human effort, while others combine model uncertainty with diverse data representation through diversity sampling. A holistic approach combines these factors with cost-effectiveness, weighting data based on anticipated reductions in loss, classification entropy, and acquisition cost. These approaches collectively aim to minimize redundancy, which occurs when a human annotates a data point that the model would predict the correct label in subsequent iterations.


In this work, we deal with the very first version of a dataset, which will serve as the foundation for iterative model improvement using AL methods. Consequently, our primary focus is not on optimizing cost-effectiveness, as the data was obtained through random sampling. Instead, we are exploring alternative data selection strategies to ensure that the initial data pool closely resembles a ``near-ideal'' random sample. This selection should not only minimize unnecessary annotations but also elevate the model's performance above the random average. To achieve this goal, we employ an uncertainty-based strategy to address two distinct challenges: identifying data points that are distant from the decision boundary and selecting examples that offer a more diverse and informative perspective on the dataset.

In addition to uncertainty estimation, various strategies are available for actively selecting data points to enhance low-resource NLP models. Diversity-based methods place their focus on achieving a balance between informativeness and the diversity of concepts or linguistic structures within the selected subset. This approach aims to prevent the model from learning biased information. Such balance can be achieved through techniques like calculating pairwise distances between data points and employing sampling strategies to select diverse examples.
For instance, \citet{sener2018active} employed the cosine similarity between word vector representations and a k-center greedy algorithm to identify the most diverse subset of data. Meanwhile, \citet{zhang2021stochastic} utilized a mutual information-based criterion to ensure that the selected data points are positioned far apart from each other in the embedding space. Additionally, there are works that combine diversity and uncertainty sampling in order to enhance the model's performance.



\section{Methods}

To tackle the challenge of determining which data to annotate, we have devised Informed Data Selection methods, which, in practice, can be thought of as ordering algorithms when executed to completion.  Random data selection can sometimes result in an imbalanced distribution of labels for human annotation, leading to an overabundance of certain labels while leaving others underrepresented. 
Our proposed methods also address this issue since the labels are not known before the annotation process. 
However, our findings indicate that our approach may be conducive to achieving a more equitable distribution of documents across various labels. 
We contend that our method is particularly well-suited for situations where humans are faced with a complete lack of labeled data.
Here, a dataset consists of words, phrases or documents that must be labeled, and will be referred to in this paper as ``documents''.

We have selected random sampling as our baseline method and have developed three additional methods for comparison against this baseline. These methods are constructed using distinct heuristics:
(i) The first method assesses semantic similarity and prioritizes documents with low similarity to those already selected; 
(ii) The second method involves clustering embeddings and systematically selects one document from each cluster based on cluster size; and 
(iii) The third method employs random sampling to choose documents with lower lexical similarity, excluding those that share too many common n-grams.
Further elaboration on these methods is provided below.

Let $\mathcal{D}$ be a set of documents $d_i$. 
Let $E$ be the set of embeddings for each document $d_i \in \mathcal{D}$. 
We define $C = \{c_1,...,c_{n_{classes}}\}$, $|C| = n_{classes}$, as the set of target classes for the classification task in supervised training.
$\mathcal{D}_{selected}$ is the set $\mathcal{D}$ rearranged  by $f$ according to the Informed DataSelection methods proposed here, with $|\mathcal{D}_{selected}| = |\mathcal{D}|$,
\begin{equation} 
    \label{eq:methods}
    \mathcal{D}_{selected} = f(n_{classes}, \mathcal{D}, E), 
\end{equation}
Elements from $\mathcal{D}_{selected}$ are then selected to constitute $D_a$ with the most relevant documents for labeling.
Let $n_{shots}$ be the target number of annotated documents per class. 
The set of annotated documents is $D_a = \{D_{a}^{c_1}, D_{a}^{c_2}, ..., D_{a}^{c_{n_{classes}}}\}$, with $|D_a| = |D_{a}^{c_1}| + |D_{a}^{c_2}| + ... + |D_{a}^{c_{n_{classes}}}|$. 
Ideally, we want $|D_{a}^{c_i}| = n_{shots}$.

The {\bf overannotation rate} $\theta$ is defined as the excess of documents annotated with the respective method used up to the target number $n_{shots}$ of annotated documents for each class $c_i \in C$, with: 
\begin{equation} 
    \label{eq:teta}
    \theta = |D_a| / (n_{classes} * n_{shots}). 
\end{equation}
It measures the excess of annotated documents generated by the method until the desired target $n_{shots}$ is achieved for each specific class $c_i$. 

We now describe the three Informed Data Selection methods proposed in this paper.

\textbf{1) Reverse Semantic Search (RSS)}: Given a set of documents $\mathcal{D}$, its respective set of embeddings $E$, and a similarity function between pairs of embeddings $sim(x_1,x_2)$, RSS calculates the similarity matrix between all embeddings of $E$. 
The similarity matrix $S$ is an $|\mathcal{D}| \times |\mathcal{D}|$ matrix whose $(i, j)$ element equals the similarity $sim(e_i, e_j)$ between $e_i, e_j \in E$, with $e_i$ and $e_j$ being the embeddings of $d_i, d_j \in \mathcal{D}$, $d_i \neq d_j$. 
RSS initially selects the two documents with the least similarity and puts both in a new set named $\mathcal{D}_{selected}$.
Then, iteratively, RSS continues to select the next most dissimilar element from the rest of the set $\{\mathcal{D} - \mathcal{D}_{selected}\}$.
RSS stops when $|\mathcal{D}_{selected}| = |\mathcal{D}|$.
In fact, RSS sorts the documents in $\mathcal{D}$ based on their dissimilarity.
The idea is that the annotation process is performed for each document in the new set generated $\mathcal{D}_{selected}$, in order, until at least $n_{shots}$ are obtained for each of the $n_{classes}$.

\textbf{2) Ordered Clustering (OC)}: Given a set of documents $\mathcal{D}$ and its respective set of embeddings $E$, OC applies a hierarchical and density-based clustering algorithm that assigns a membership probability to each document in relation to each cluster, indicating the probability of that document being in that cluster. Then, OC orders the clusters based on their size, i.e., based on the number of documents that belong to a given cluster. 
Finally, OC exhaustively selects the document with the lowest membership probability from each cluster, from largest to smallest cluster, and removes it from the cluster, placing it, in removal order, in $\mathcal{D}_{selected}$.
The OC iterative process stops when all clusters are empty. 
Here too, the annotation process is performed for each document in the new set generated $\mathcal{D}_{selected}$, in order, until at least $n_{shots}$ are obtained for each of the $n_{classes}$.

\textbf{3) Limited Lexical Similarity (LLS)}: Given a set of documents $\mathcal{D}$, a lexical comparison function $g(d_1,d_2)$ (based on BLEU score, ROUGE score or other metrics) and a threshold value $\beta$, LLS chooses the first document $d_i$ randomly and inserts it into the initially empty set $\mathcal{D}_{selected}$. 
LLS then proceeds by choosing the next document $d_{i+1}$ at random, discarding it if $g(d_{i+1},d_{i}) > \beta$ and keeping it otherwise. LLS stops when there are no more documents to select.
Similar to the RSS and OC methods, the generated set can have many elements. 
Note that in this case, $|\mathcal{D}_{selected}|$ may be smaller than $|\mathcal{D}|$, given that some documents were discarded.
Thus, the annotation takes place by removing documents from $\mathcal{D}_{selected}$, in the order in which they were inserted in $\mathcal{D}_{selected}$, until at least $n_{shots}$ are obtained for each of the $n_{classes}$.

\section{Experimental Setup}

\begin{figure*}
    \centering
    \footnotesize
    \includegraphics[scale=0.55]{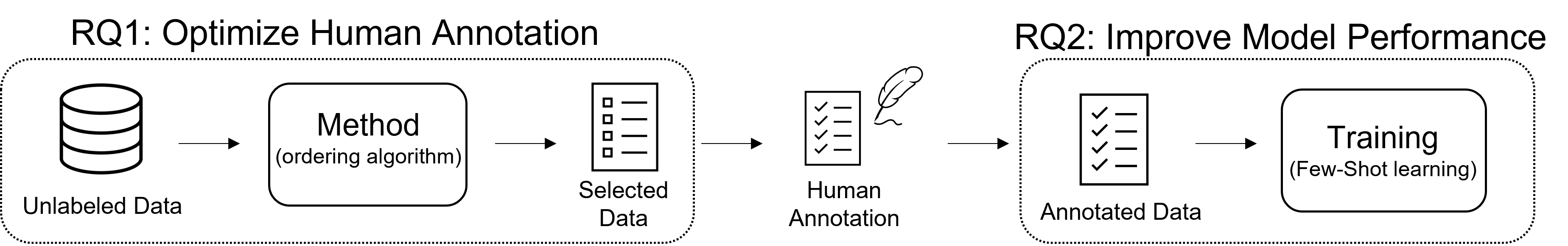}
  \caption{Full Architecture of our Settings. Results from RQ 1 are evaluated with metric Overannotation Rate. Results from RQ 2 use metrics Accuracy and Macro-F1 Score.}
\end{figure*}

This section outlines the experimental setup for evaluating our proposed Informed Data Selection architecture. The evaluation is conducted on five text classification datasets, selected to explore varying degrees of data imbalance, class diversity, language, and domain. In this section, we present the datasets used and describe two key experimental settings: \textit{Human Annotation} and \textit{Few-shot learning with selected data}.

\subsection{Datasets}

We use the following datasets in our experiments:

\begin{itemize}
\item \textbf{AgNews} \citep{agnews_ref}: A news dataset with 4 classes and balanced data distribution. It consists of 120,000 training examples and 7,600 test examples, available only in English.

\item \textbf{SST5} \citep{sst5_ref}: A sentiment analysis dataset with 5 classes and a slightly imbalanced data distribution. It contains 8,544 training examples, 1,101 validation examples, and 2,210 test examples, available in English.

\item \textbf{Emotion} \citep{emotion_ref}: An emotion analysis dataset with 5 classes and imbalanced data distribution. It includes 16,000 training examples and 2,000 test examples, available in English.

\item \textbf{Multilingual Sentiment Analysis (MSA)} \footnote{Available on \url{https://huggingface.co/datasets/tyqiangz/multilingual-sentiments}}: A multilingual sentiment analysis dataset with 3 classes and balanced data distribution. We make use of the Portuguese subset of this dataset, that contains 1,839 training examples and 870 test examples. 

\item \textbf{BRNews} \footnote{Available on \url{https://huggingface.co/datasets/iara-project/news-articles-ptbr-dataset}}: A Brazilian Portuguese news dataset with 19 classes and imbalanced data distribution. It comprises 176,114 training examples and 176,114 test examples, available only in Portuguese.
\end{itemize}

The train and test splits are utilized for training and evaluation, unless specified otherwise. An overview of these datasets is provided in Table 
\ref{tab:dataset-characteristics}. The choice of these datasets aims at isolating and scrutinizing key data distribution variables. Our focus centers on examining the impact of factors such as the number of samples per class, the quantity of classes within each dataset, the extent of data imbalance, and the language (English or Portuguese) on the outcomes of Informed Data Selection methods.

\begin{table}[ht]
\caption{Datasets Characteristics}
\label{tab:dataset-characteristics}
\resizebox{\columnwidth}{!}{
\begin{tabular}{|l|l|l|l|l|}
\hline
{\bf Dataset} & {\bf \# docs} & {\bf classes} & {\bf Balancing }          & {\bf Lang }  \\ \hline
AgNews        &    127600     &     4         & balanced            & En    \\ \hline
SST5         &    11855      &     5         & slightly imbalanced & En    \\ \hline
Emotion       &    18000      &     6         & imbalanced          & En    \\ \hline
MSA &    3033       &     3         & balanced            & Pt \\ \hline
BRNews         &    352228     &    19         & very imbalanced     & Pt \\ \hline
\end{tabular}
}
\end{table}

\subsection{Research Questions}

In our study, we aim to address specific research questions through distinct experimental settings, each designed to provide insights into the efficacy of our Informed Data Selection methods. These experimental settings are detailed below.

\subsubsection{RQ1: Which method allows for more efficient human annotation?}

To tackle this question, we simulate a real-life scenario where no annotated data is initially available, and human annotators are required to annotate the data. We compare different sorting methods designed to prioritize annotation and, leveraging known ground-truth, we quantify the overannotation rate (see Eq. \ref{eq:methods}) that each method might entail. In this context, we compare the performance of our Informed Data Selection methods with that of a random sampling strategy, referred to as \textbf{Random}.

\subsubsection{RQ2: Which method yields better few-shot learning?}

To address this second question, we turn our attention to models trained on the dataset created in the context of RQ1. The goal is to determine whether the more efficient annotation process comes with a price, and could potentially lead to biased models, resulting in decreased performance compared to conventional random sampling. Conversely, our initial hypothesis suggests that Informed Data Selection, by increasing data diversity, will lead to model improvement, as it provides more knowledge with same amount of training data.


\subsection{Evaluation Metrics}

Within the context of RQ1 setting, the primary evaluation metric is the {\bf overannotation rate} $\theta$ (Eq. \ref{eq:teta}). 
This metric is relevant as in resource-constrained scenarios, the imperative lies in the minimization of excessive annotation. For this metric \textbf{lower values} mean more efficiency.

As for the RQ2 setting, we employ conventional metrics commonly used in text classification. These include \textbf{Accuracy}, which measures the percentage of correctly classified instances, and, exclusively for the very imbalanced dataset, the \textbf{Macro F1-score}, a metric that calculates the harmonic mean of precision and recall for each class and then averages these values across all classes.

\subsection{Implementation Details}

For addressing RQ1, our chosen embedding model for RSS and OC is \texttt{paraphrase-multilingual-mpnet-base-v2}\footnote{Available on \url{https://huggingface.co/sentence-transformers/paraphrase-mpnet-base-v2}}. To perform clustering in OC, we employ the HDBSCAN algorithm \citep{campello2013hdbscan}. We employ BLEU score \citep{papineni2002bleu} as comparison function in LLS. The entire process for LLS and Random is executed identically 10 times, and results are reported as mean values along with confidence intervals. 

Regarding RQ2, we train models under two distinct configurations to isolate the influence of the training algorithm for few-shot learning. We utilize the HuggingFace Transformers library \citep{wolf2020transformers} and employ the following methods:

\begin{itemize}
\item \textsc{FineTune}: We fine-tune the \texttt{XLM-Roberta-large} \citep{conneau-etal-2020-unsupervised}, a pre-trained encoder-based Language Model, following conventional fine-tuning procedures for Sequence Classification. The training process spans 30 epochs with a learning rate of $2\times10^{-5}$. 

\item \textsc{SetFit}: For this method, we utilize Sentence Transform fine-tuning (SetFit) \citep{setfit}, an efficient approach for few-shot learning in encoder-based models. SetFit dynamically generates training pairs from annotated data and leverages contrastive loss for training the model on the classification task. As the base model, we also use \texttt{paraphrase-multilingual-mpnet-base-v2}.
\end{itemize}

Results in RQ2 for LLS and Random, which exhibit stochastic behavior, are presented in terms of mean values and standard deviations across 10 runs. The experiments are conducted across a range of $n_{shots}$ values, specifically 8, 16, 32, and 64, with a batch size of 16 for the training process.





%

\section{Results}
\begin{figure}
\centering
  \begin{minipage}{0.32\textwidth}
    \centering
    \footnotesize
    $\quad$AgNews
    \includegraphics[width=\linewidth]{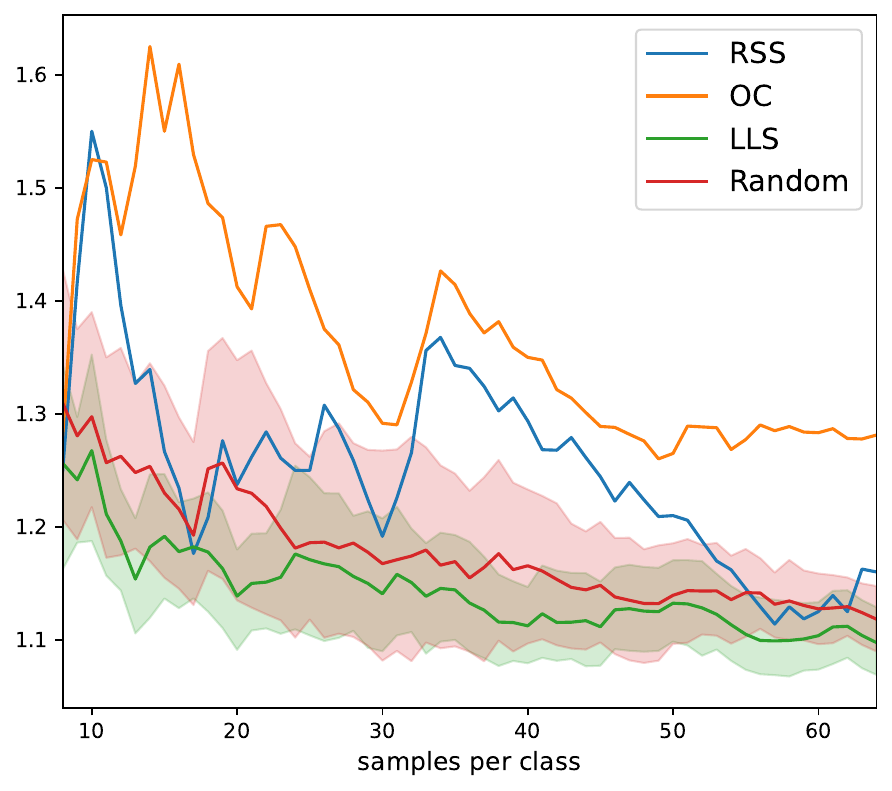}
  \end{minipage}%
  \vfill
  \begin{minipage}{0.32\textwidth}
    \centering
    \footnotesize
    $\quad$MSA
    \includegraphics[width=\linewidth]{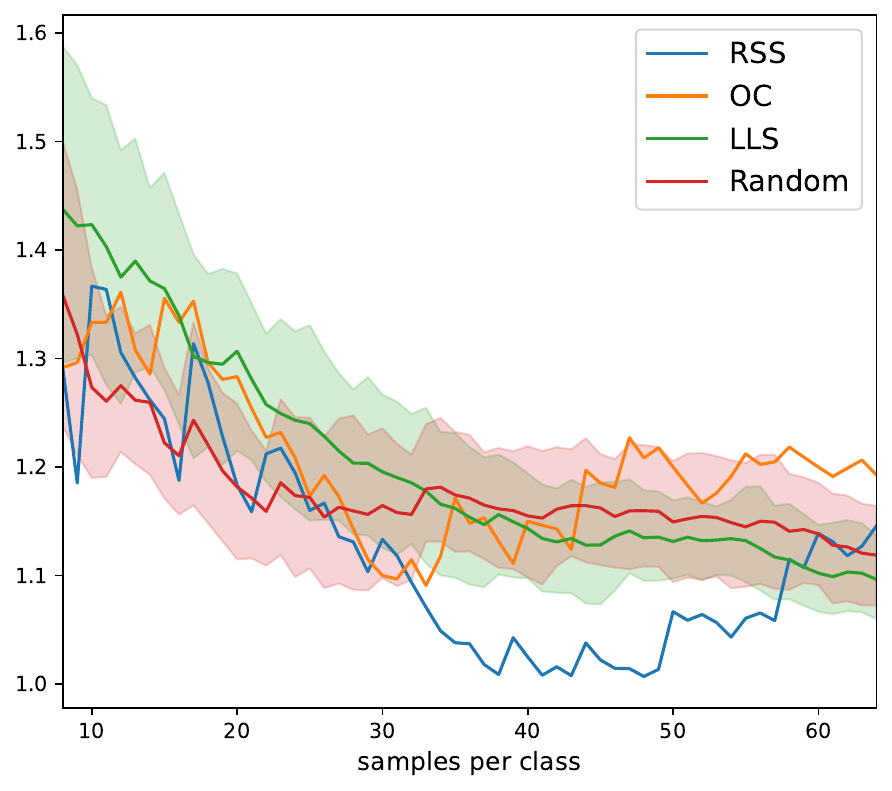}
  \end{minipage}%
  \vfill
  \begin{minipage}{0.32\textwidth}
    \centering
    \footnotesize
    $\quad$SST5
    \includegraphics[width=\linewidth]{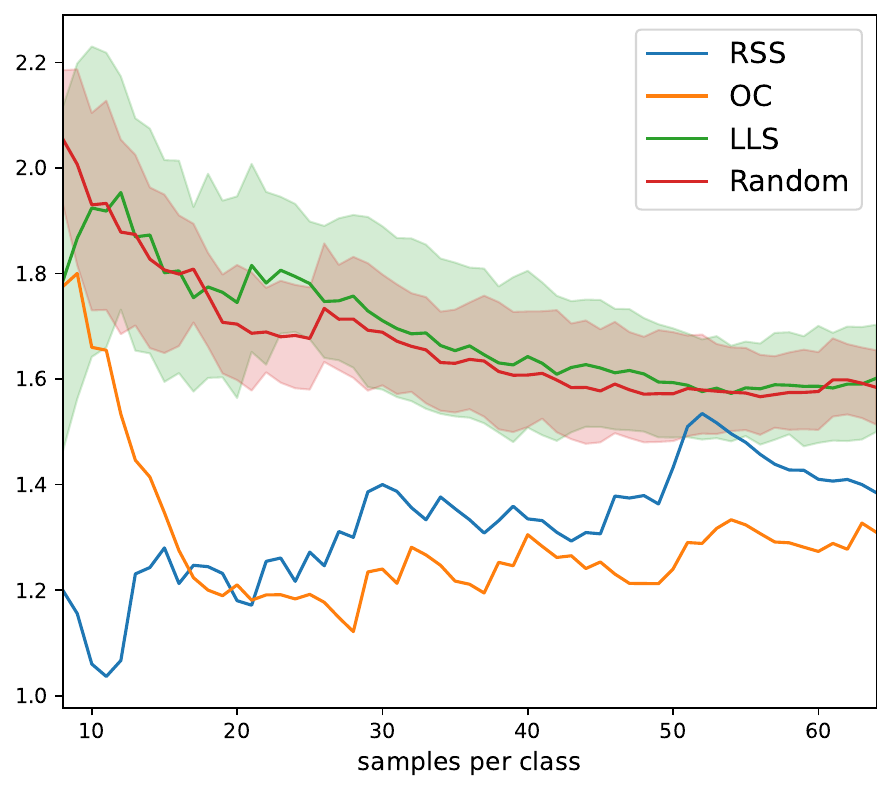}
  \end{minipage}%
  \vfill
  \begin{minipage}{0.32\textwidth}
    \centering
    \footnotesize
    $\quad$Emotion
    \includegraphics[width=\linewidth]{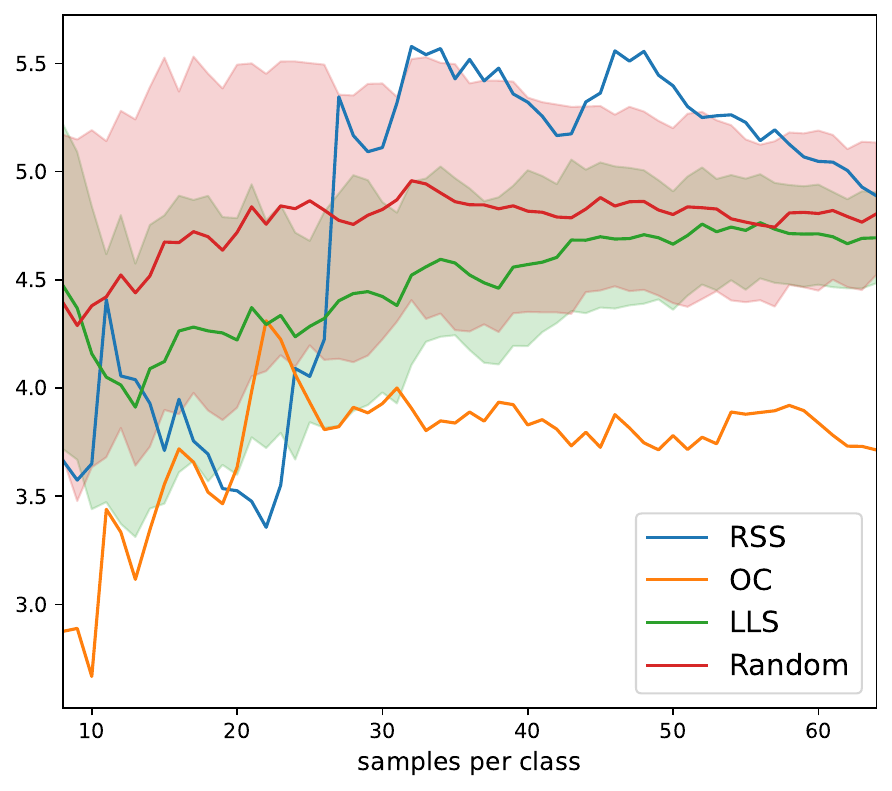}
  \end{minipage}
    \begin{minipage}{0.32\textwidth}
    \centering
    \footnotesize
    $\quad$BRNews\\
    \includegraphics[width=\linewidth]{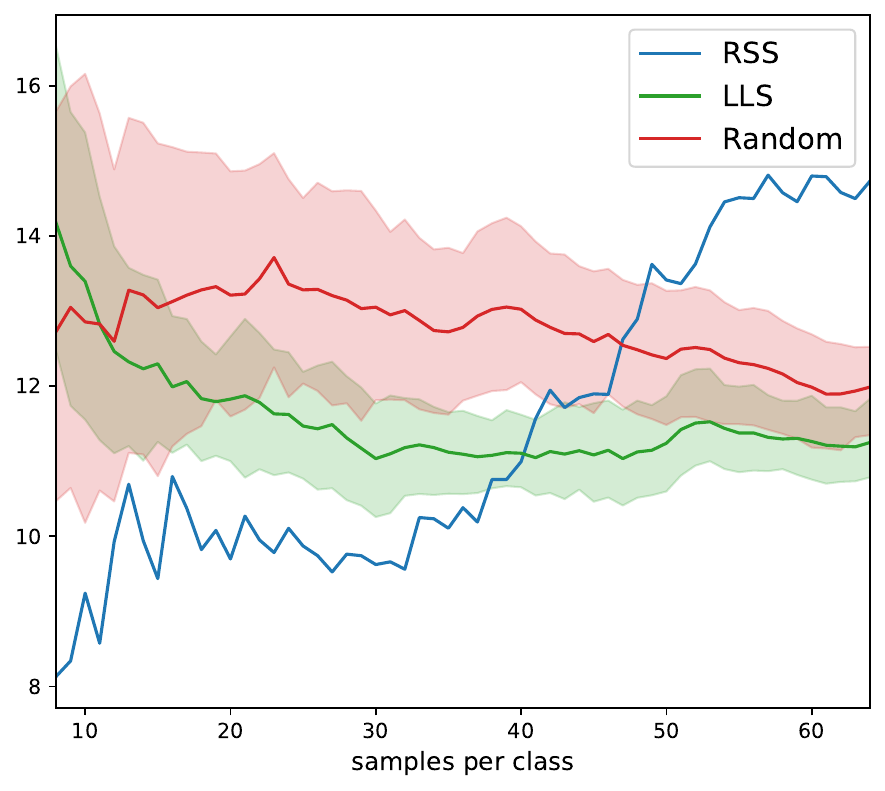}
  \end{minipage}
  \caption{Overannotation Rate $\theta$ per Dataset and Method.}
\label{fig:rq1}
\end{figure}

We compare the performance of our proposed Informed Data Selection methods with random sampling strategy on the five datasets. 

\subsection{Efficiency in Human Annotation (RQ1)}

Charts in Figure \ref{fig:rq1} show results for experiments where we measure the overannotation rate $\theta$ as a function of the number of samples per class in the Dataset, for each method (RSS, OC, LLS and Random as baseline). 
Methods LLS and Random are executed 10 times, and averaged results (along with confidence interval) are shown. 

In \textbf{balanced datasets}, we observe that \textbf{no method consistently outperforms} the random baseline. This is seen in AgNews and MSA. It can be explained, as we have mentioned before, by the distribution of classes in these datasets: both are heavily balanced, which tend to favor random sampling methods. So, when it comes to balanced data distributions, the human may not worry about overannotation of the random method. It is interesting to note that in MSA, when $n_{shots}$ is in the range of 30 to 60, RSS would indeed be a better choice than random sampling. Also, our methods are slightly more competent in MSA than in AgNews. The language factor may play a minor role here: because our embedding model, although multilingual, was trained on more English than Portuguese data, its embeddings are less tuned to the Portuguese language, which might explain why RSS promotes variety for a longer range of $n_{shots}$, but eventually converges with most other methods. Aside from this possible model-related factor, language does not seem to be a relevant factor for our selection methods.

For \textbf{imbalanced data distributions}, two of our methods consistently outperform random sampling: RSS and OC. We observe a lower overannotation rate $\theta$ in SST5 and Emotion when $n_{shots} < 30$, indicating that both RSS and OC are a better fit than random sampling in imbalanced distributions. As we increase $n_{shots}$ further from 30, only RSS in the Emotion dataset worsens, but methods are overall are more efficient in choosing which data to annotate, generating less excess of annotations.


For a \textbf{heavily imbalanced distribution}, we see a different behavior. We observe that as \textbf{number of classes and data imbalance grow, overannotation rate $\theta$ increases} for every method tested (BRNews has 10 times more overannotation rate than balanced datasets). In turn, OC generates too much overannotation rate (more than 6 times than Random baseline), and is thus considered an outlier and excluded from the chart. 

Results show that \textbf{RSS considerably outperforms Random baseline} for $n_{shots} < 40$. This is once again due to the fact that this dataset has a much higher number of classes, with very imbalanced distribution of documents per each class, much closer to a real-life scenario humans find themselves. 
In these scenarios, our method thrives, generating as few as half excess annotations when compared to the Random method. However, as observed for every dataset, our methods and Random baseline also converge when $n_{shots}$ increases further away from around 50. 

\subsection{Model Performance (RQ2)}

\begin{figure}
  \caption*{AgNews}
  \begin{minipage}{0.25\textwidth}
    \centering
    \footnotesize
    $\quad$\textsc{FineTune}
    \includegraphics[width=\linewidth]{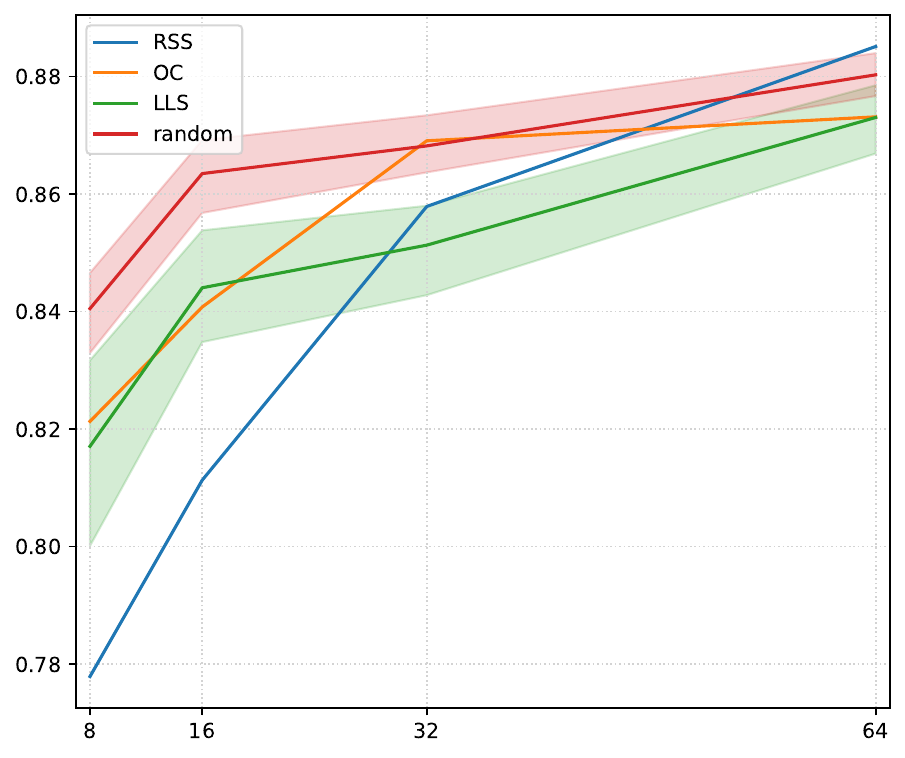}
  \end{minipage}%
  \begin{minipage}{0.25\textwidth}
    \centering
    \footnotesize
    $\quad$\textsc{SetFit}
    \includegraphics[width=\linewidth]{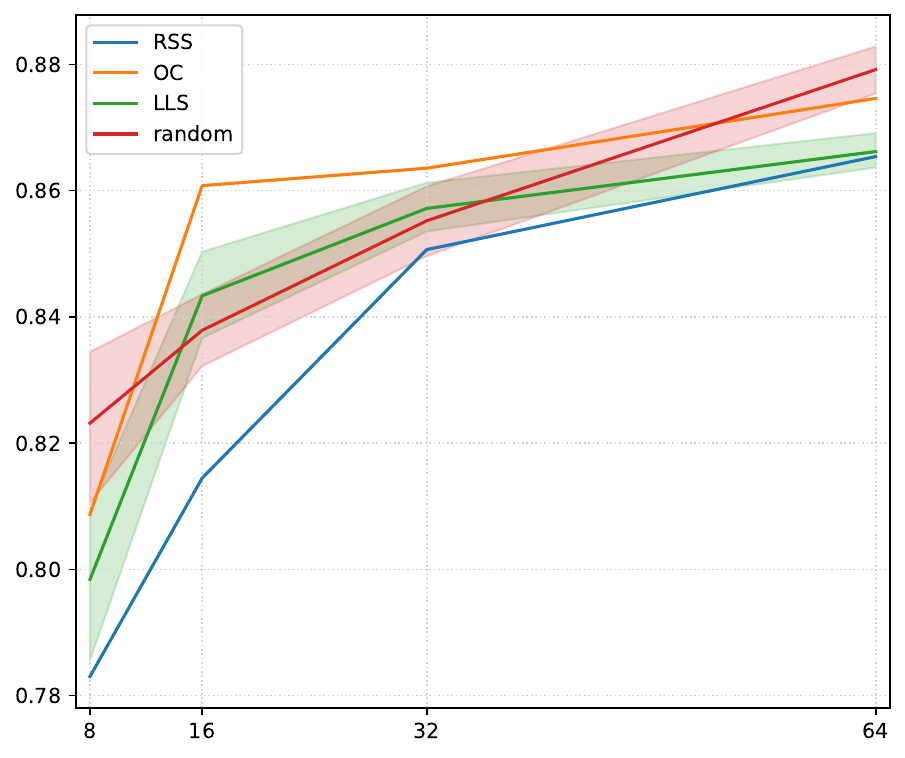}
  \end{minipage}%
  \vfill
    \caption*{SST5}
   \begin{minipage}{0.25\textwidth}
    \centering
    \footnotesize
        $\quad$\textsc{FineTune}

    \includegraphics[width=\linewidth]{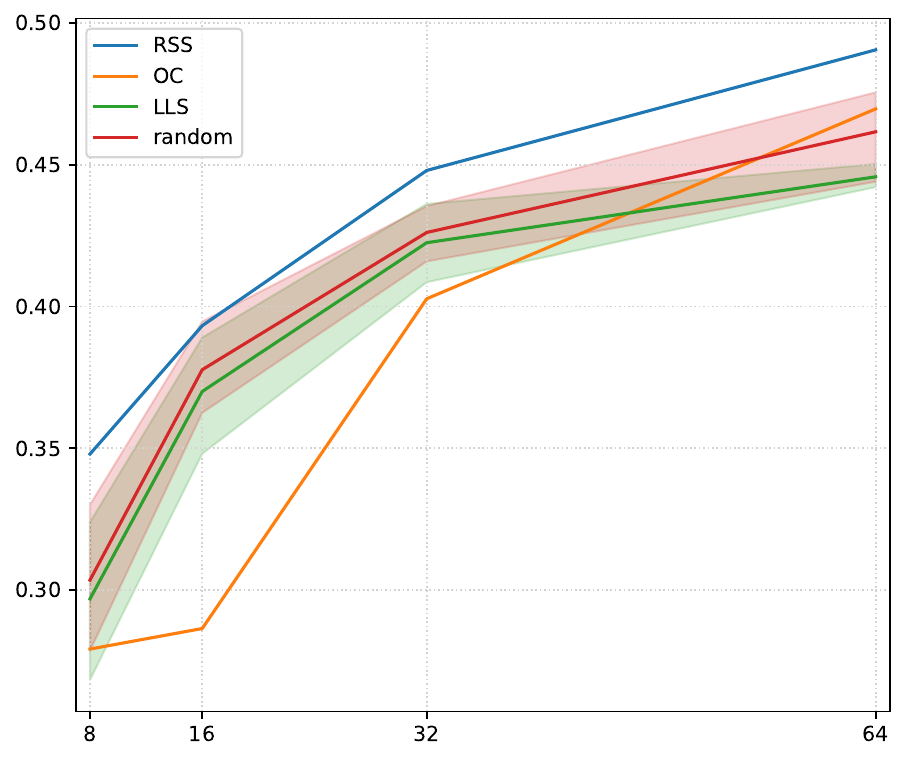}
  \end{minipage}%
  \begin{minipage}{0.25\textwidth}
    \centering
    \footnotesize
        $\quad$\textsc{SetFit}

    \includegraphics[width=\linewidth]{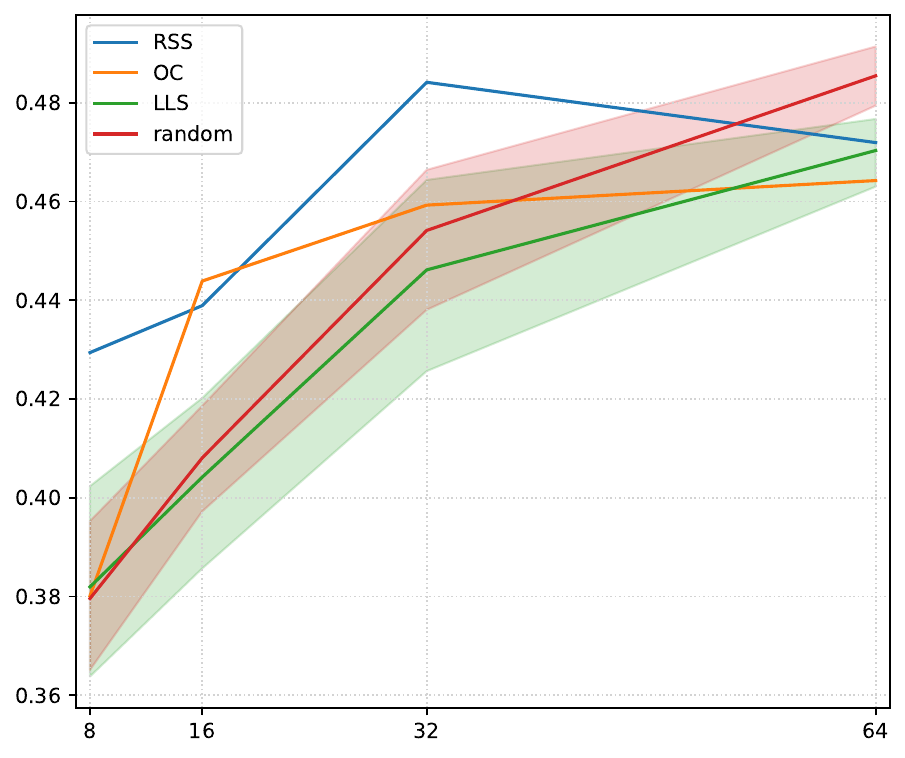}
  \end{minipage}%
  \vfill
  \caption*{Emotion}
   \begin{minipage}{0.25\textwidth}
    \centering
    \footnotesize
            $\quad$\textsc{FineTune}
    \includegraphics[width=\linewidth]{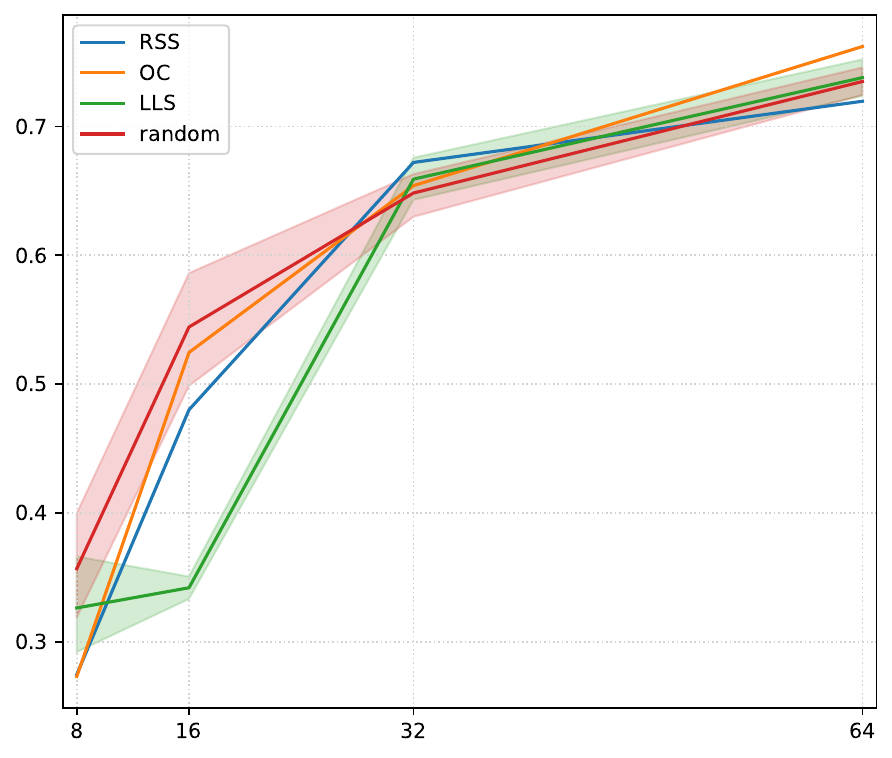}
  \end{minipage}%
  \begin{minipage}{0.25\textwidth}
    \centering
    \footnotesize
        $\quad$\textsc{SetFit}
    \includegraphics[width=\linewidth]{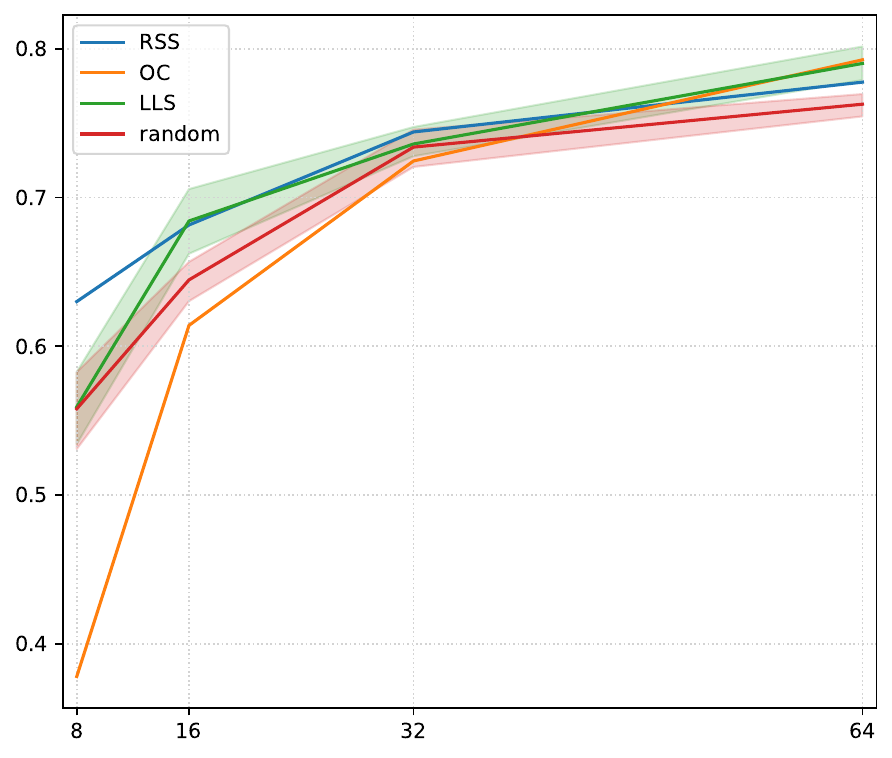}
  \end{minipage}%
  \vfill
    \caption*{MSA}
   \begin{minipage}{0.25\textwidth}
    \centering
    \footnotesize
                $\quad$\textsc{FineTune}
    \includegraphics[width=\linewidth]{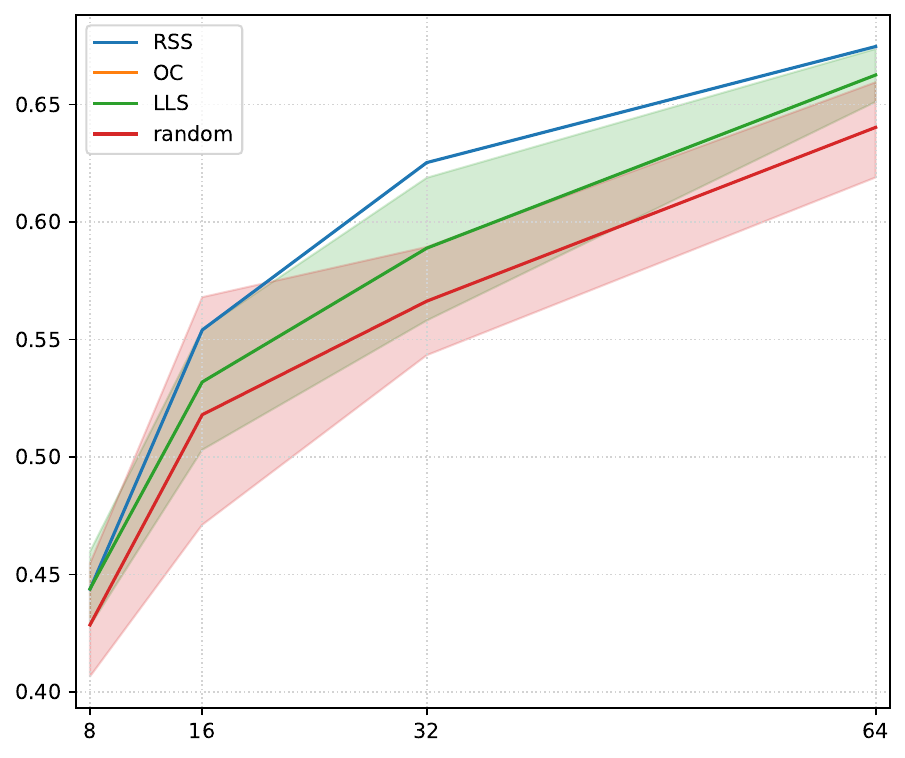}
  \end{minipage}%
  \begin{minipage}{0.25\textwidth}
    \centering
    \footnotesize
        $\quad$\textsc{SetFit}
    \includegraphics[width=\linewidth]{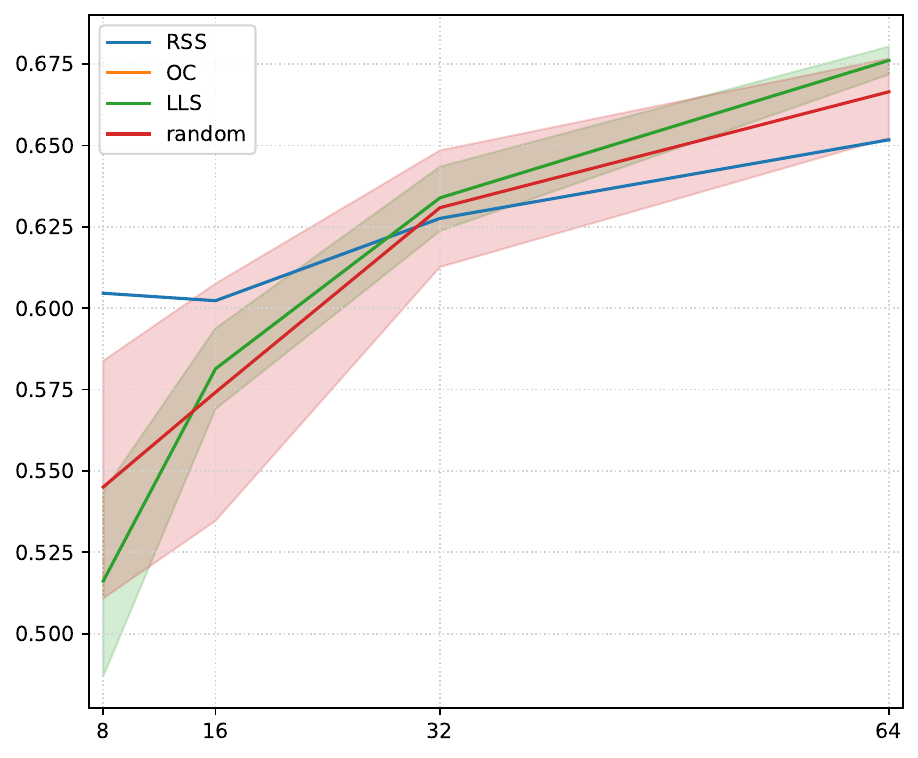}
  \end{minipage}%
  \vfill
      \caption*{BR News}
   \begin{minipage}{0.25\textwidth}
    \centering
    \footnotesize
            $\quad$\textsc{FineTune}
    \includegraphics[width=\linewidth]{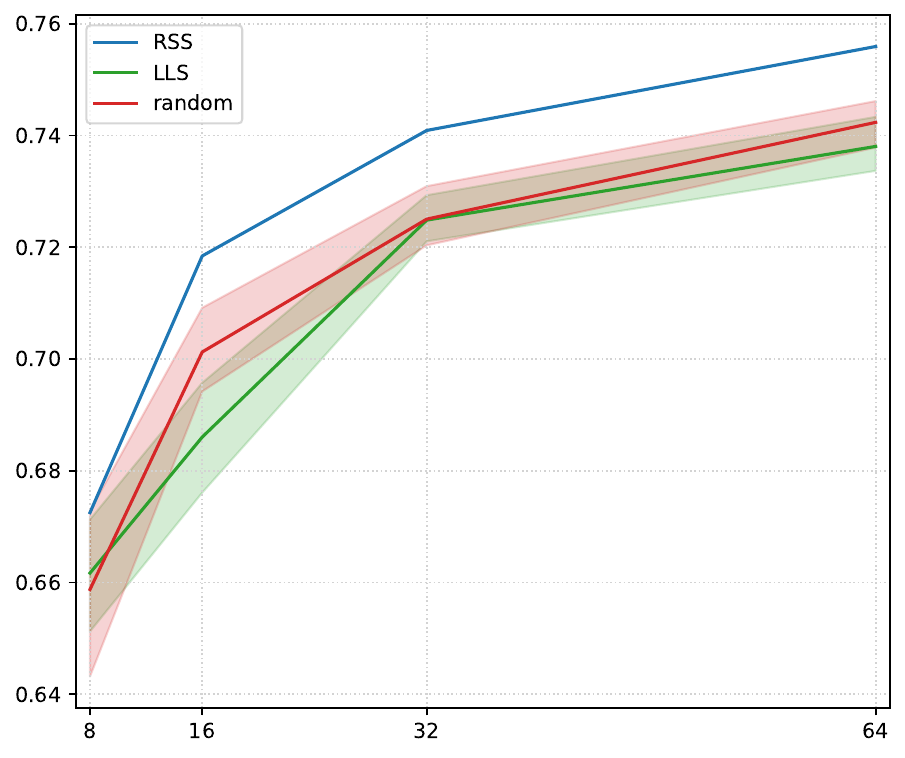}
  \end{minipage}%
  \begin{minipage}{0.25\textwidth}
    \centering
    \footnotesize
        $\quad$\textsc{SetFit}
    \includegraphics[width=\linewidth]{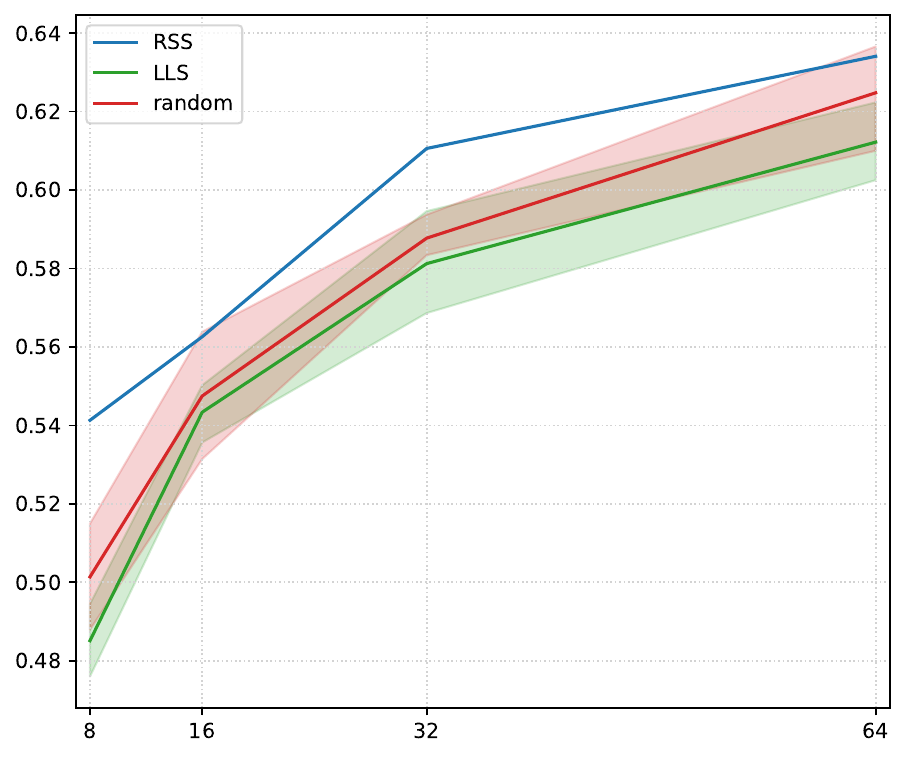}
  \end{minipage}%
  \caption{Accuracy over evaluation datasets.}
  \label{fig:rq2}
\end{figure}

Figure \ref{fig:rq2} shows results for experiments where we compare the performance of classifiers trained with data selected by our methods and the Random method. Because OC fails to generate a feasible excess of annotation for BRNews, it is deemed as not applicable and therefore excluded from reports. 

As a general result, we observe that our methods \textbf{OC and LLS fail to consistently outperform} the Random baseline. However, \textbf{RSS outperforms random sampling in almost every scenario}. For both \textsc{FineTune} and \textsc{SetFit}, RSS is better than random sampling for every dataset with the exception of the AgNews, where random sampling yields higher accuracy. A mix of many factors may be responsible for this: first, AgNews is balanced, which favors random sampling when selecting training data; second, the task of AgNews is simple when compared to other datasets, because classes in it have distinct traits (ie. they refer to distinct themes, such as Sports, Technology, etc) which may help with decision boundaries of the model. The other balanced dataset, MSA, does not have these distinct traits for its classes, which instead express a kind of gradation (ie. Positive, Neutral, Negative). In other words, the classification task in MSA is tougher, which means that selecting data with more variability can effectively boost model performance.

We note that \textbf{the higher the degree of data imbalance, the more consistently RSS will outperform random sampling}. However, reporting only accuracy in a heavily imbalanced dataset is insufficient to adequately represent performance of a classifier. Thus, Table \ref{tab:macro-f1-brnews} shows results of \textbf{Macro-F1 Score} for both training methods in BRNews. We see that for \textsc{FineTune},\textbf{ RSS performs consistently better}, while \textsc{SetFit} also shows a slight improvement when compared to Random, falling above the confidence interval only for $n_{shots}$ = 8. This is an indicative that \textbf{both RSS and Random methods perform almost equally well across classes}, disconsidering imbalance among classes. This indicates that both methods succeed at selecting diverse data for model training. Still, \textbf{RSS provides higher accuracy}, turning it into the \textbf{recommended method for low-resource setups}.

\begin{table}[ht]
\caption{FineTune and SetFit F1-Score Macro for Random Selection on BRNews dataset.}
\label{tab:macro-f1-brnews}
\small
\centering
\begin{tabular}{|l|c|c|c|}
\hline
\textbf{Training} & $n_{shots}$ & \textbf{RSS} & \textbf{Random} \\
\hline
\textsc{FineTune} & 8 & $58.6$  & $56.7 \pm 2.3$ \\
\textsc{FineTune} & 16 & $62.0$ & $60.6 \pm 0.8$ \\
\textsc{FineTune} & 32 & $65.2$ & $62.8 \pm 0.9$ \\
\textsc{FineTune} & 64 & $66.8$ & $63.6 \pm 0.5$ \\
\textsc{SetFit}   & 8 & $46.83$ & $45.0 \pm 1.7$ \\
\textsc{SetFit}   & 16 & $48.8$ & $48.8 \pm 1.9$ \\
\textsc{SetFit}   & 32 & $52.3$ & $52.2 \pm 1.0$ \\
\textsc{SetFit}   & 64 & $55.9$ & $55.6 \pm 1.2$ \\
\hline
\end{tabular}
\end{table}

Another important result is \textbf{the convergence of all methods when $n_{shots}$ grows}. Because our methods are suited to the construction of a very first version of a dataset for Active Learning, both overannotation rate and model performance converge when $n_{shots} > 64$. A reason is that, as the number of selected data grows, diversity will also grow. Although results show that our selection methods promote more diversity for lower $n_{shots}$, any selection method that does not apply oversampling will bring diversity if $n_{shots}$ keeps increasing. Thus, other methods outpace ours in promoting diversity when we leave the realm of few-shot -- i.e. when we annotate too much data. This means that \textbf{when the desire is to annotate lots of data per class, most methods evaluated in this work are not suited to the selection}, with random sampling being a better strategy.

\section{Conclusion}

This work has proposed an automatic Informed Data Selection architecture which aims to select which data should be annotated by a human to build a first dataset. We simulated 2 scenarios, and experimental results we report show our architecture is a better option than random sampling methods for few-shot learning. We have shown that the higher the imbalance in the dataset, the more competent our method is -- both in generating less excess of annotation and in improving model performance. As far as we know, there are few works that address the imbalance problem as a variable of a supervised dataset.

In particular, the Reverse Semantic Search (RSS) method has shown to be the most competent in experiments across different languages, number and imbalance across classes. However, it should be noted that for Limited Lexical Similarity (LLS), a numeric threshold is specified. This work has not fine-tuned this hyperparameter, instead it was chosen by manually inspecting results with different thresholds. 

Results indicate that fine-tuning this threshold can improve the overannotation rate and accuracy of LLS, as its standard deviation in many datasets is smaller than that of the Random method. The same can be said about using another comparison function, such as ROUGE score. The need for fine-tuning may be amplified in very specialized domains with unusual vocabulary -- however, more experiments are needed to confirm this observation.

We also note that the Ordered Clustering (OC) method did not provide consistent results across many datasets. Because our method relies on picking one document from each cluster, when the number of identified clusters is high, OC fail to select quality data. This can be addressed by combining clustering with RSS or LLS, and is an attractive direction for future work.


Our concluding remark underscores a frequently overlooked aspect in the realm of few-shot learning, particularly in scenarios where labeled data is scarce. Many studies in few-shot learning often assess their methods across a range of datasets, typically characterized by a balance or slight imbalance in class distribution. 
While there are exceptions, those works that do evaluate on imbalanced datasets often fail to adequately address the consequences of such imbalance. 
The reality is that in real-world applications, balanced data distributions are a rarity. 
Hence, we advocate that authors engaged in few-shot learning techniques should be cognizant of this reality, and whenever feasible, report metrics that account for imbalance in their evaluations.

\section*{Acknowledgements}
This work is financed by National Funds through the Portuguese funding agency, FCT - Fundação para a Ciência e a Tecnologia, within project LA/P/0063/2020.
DOI 10.54499/LA/P/0063/2020 | https://doi.org/10.54499/LA/P/0063/2020.
This work is also supported in part by the Brazilian National Council for Scientific and Technological Development (CNPq) under grant numbers 310085/2020-9, and the Brazilian Coordination for the Improvement of Higher Education Personnel (CAPES, Finance Code 001). Israel Campos Fama is funded by the Secretaria de Fazenda do Estado do Rio Grande do Sul (Sefaz-RS). Bárbara Dias Bueno is funded by the \textit{Programa Unificado de Bolsas (PUB)} undergraduate research scholarship from Universidade de São Paulo, under project 2117/2023. 


\bibliography{anthology,custom}

\begin{thebibliography}{29}
\expandafter\ifx\csname natexlab\endcsname\relax\def\natexlab#1{#1}\fi

\bibitem[{Alcoforado et~al.(2022)Alcoforado, Ferraz, Gerber, Bustos, Oliveira, Veloso, Siqueira, and Costa}]{alcoforado2022zeroberto}
Alexandre Alcoforado, Thomas~Palmeira Ferraz, Rodrigo Gerber, Enzo Bustos, Andr{\'e}~Seidel Oliveira, Bruno~Miguel Veloso, Fabio~Levy Siqueira, and Anna Helena~Reali Costa. 2022.
\newblock \href {https://arxiv.org/abs/2201.01337} {{ZeroBERTo: Leveraging Zero-Shot Text Classification by Topic Modeling}}.
\newblock In \emph{International Conference on Computational Processing of the Portuguese Language (PROPOR 2022)}, pages 125--136. Springer.

\bibitem[{Basile et~al.(2021)Basile, P{\'e}rez-Torr{\'o}, and Franco-Salvador}]{basile2021probabilistic}
Angelo Basile, Guillermo P{\'e}rez-Torr{\'o}, and Marc Franco-Salvador. 2021.
\newblock \href {https://aclanthology.org/2021.ranlp-1.16} {{Probabilistic Ensembles of Zero- and Few-Shot Learning Models for Emotion Classification}}.
\newblock In \emph{Proceedings of the International Conference on Recent Advances in Natural Language Processing (RANLP 2021)}, pages 128--137, Held Online.

\bibitem[{Beijbom(2014)}]{beijbom2014random}
Oscar Beijbom. 2014.
\newblock \href {https://arxiv.org/abs/1410.7074} {{Random Sampling in an Age of Automation: Minimizing Expenditures through Balanced Collection and Annotation}}.
\newblock \emph{arXiv preprint arXiv:1410.7074}.

\bibitem[{Brown et~al.(2020)Brown, Mann, Ryder, Subbiah, Kaplan, Dhariwal, Neelakantan, Shyam, Sastry, Askell, Agarwal, Herbert-Voss, Krueger, Henighan, Child, Ramesh, Ziegler, Wu, Winter, Hesse, Chen, Sigler, Litwin, Gray, Chess, Clark, Berner, McCandlish, Radford, Sutskever, and Amodei}]{brown2020language}
Tom Brown, Benjamin Mann, Nick Ryder, Melanie Subbiah, Jared~D Kaplan, Prafulla Dhariwal, Arvind Neelakantan, Pranav Shyam, Girish Sastry, Amanda Askell, Sandhini Agarwal, Ariel Herbert-Voss, Gretchen Krueger, Tom Henighan, Rewon Child, Aditya Ramesh, Daniel Ziegler, Jeffrey Wu, Clemens Winter, Chris Hesse, Mark Chen, Eric Sigler, Mateusz Litwin, Scott Gray, Benjamin Chess, Jack Clark, Christopher Berner, Sam McCandlish, Alec Radford, Ilya Sutskever, and Dario Amodei. 2020.
\newblock \href {https://proceedings.neurips.cc/paper_files/paper/2020/file/1457c0d6bfcb4967418bfb8ac142f64a-Paper.pdf} {{Language Models are Few-Shot Learners}}.
\newblock In \emph{Advances in Neural Information Processing Systems}, volume~33, pages 1877--1901.

\bibitem[{Campello et~al.(2013)Campello, Moulavi, and Sander}]{campello2013hdbscan}
Ricardo J. G.~B. Campello, Davoud Moulavi, and Joerg Sander. 2013.
\newblock \href {https://link.springer.com/chapter/10.1007/978-3-642-37456-2_14} {Density-based clustering based on hierarchical density estimates}.
\newblock In \emph{Advances in Knowledge Discovery and Data Mining}, pages 160--172, Berlin, Heidelberg. Springer Berlin Heidelberg.

\bibitem[{Conneau et~al.(2020)Conneau, Khandelwal, Goyal, Chaudhary, Wenzek, Guzm{\'a}n, Grave, Ott, Zettlemoyer, and Stoyanov}]{conneau-etal-2020-unsupervised}
Alexis Conneau, Kartikay Khandelwal, Naman Goyal, Vishrav Chaudhary, Guillaume Wenzek, Francisco Guzm{\'a}n, Edouard Grave, Myle Ott, Luke Zettlemoyer, and Veselin Stoyanov. 2020.
\newblock \href {https://doi.org/10.18653/v1/2020.acl-main.747} {Unsupervised cross-lingual representation learning at scale}.
\newblock In \emph{Proceedings of the 58th Annual Meeting of the Association for Computational Linguistics}, pages 8440--8451, Online. Association for Computational Linguistics.

\bibitem[{Ducoffe and Precioso(2018)}]{ducoffe2018adversarial}
Melanie Ducoffe and Frederic Precioso. 2018.
\newblock \href {https://arxiv.org/abs/1802.09841} {{Adversarial Active Learning for Deep Networks: a Margin Based Approach}}.
\newblock \emph{arXiv preprint arXiv:1802.09841}.

\bibitem[{Ferraz et~al.(2021)Ferraz, Alcoforado, Bustos, Oliveira, Gerber, M{\"u}ller, d’Almeida, Veloso, and Costa}]{ferraz2021debacer}
Thomas~Palmeira Ferraz, Alexandre Alcoforado, Enzo Bustos, Andr{\'e}~Seidel Oliveira, Rodrigo Gerber, Na{\'\i}de M{\"u}ller, Andr{\'e}~Corr{\^e}a d’Almeida, Bruno~Miguel Veloso, and Anna Helena~Reali Costa. 2021.
\newblock \href {https://arxiv.org/abs/2112.05438} {{DEBACER: a method for slicing moderated debates}}.
\newblock In \emph{Anais do XVIII Encontro Nacional de Intelig{\^e}ncia Artificial e Computacional}, pages 667--678. SBC.

\bibitem[{Ferraz et~al.(2023)Ferraz, Boito, Brun, and Nikoulina}]{ferraz2023distilwhisper}
Thomas~Palmeira Ferraz, Marcely~Zanon Boito, Caroline Brun, and Vassilina Nikoulina. 2023.
\newblock \href {https://arxiv.org/abs/2311.01070} {{Multilingual DistilWhisper: Efficient Distillation of Multi-task Speech Models via Language-Specific Experts}}.
\newblock \emph{arXiv preprint arXiv:2311.01070}.

\bibitem[{Hovy et~al.(2013)Hovy, Berg-Kirkpatrick, Vaswani, and Hovy}]{hovy2013learning}
Dirk Hovy, Taylor Berg-Kirkpatrick, Ashish Vaswani, and Eduard Hovy. 2013.
\newblock \href {https://aclanthology.org/N13-1132} {Learning whom to trust with {MACE}}.
\newblock In \emph{Proceedings of the 2013 Conference of the North {A}merican Chapter of the Association for Computational Linguistics: Human Language Technologies}, pages 1120--1130, Atlanta, Georgia. Association for Computational Linguistics.

\bibitem[{Hsueh et~al.(2009)Hsueh, Melville, and Sindhwani}]{hsueh2009data}
Pei-Yun Hsueh, Prem Melville, and Vikas Sindhwani. 2009.
\newblock \href {https://aclanthology.org/W09-1904} {Data quality from crowdsourcing: A study of annotation selection criteria}.
\newblock In \emph{Proceedings of the {NAACL} {HLT} 2009 Workshop on Active Learning for Natural Language Processing}, pages 27--35, Boulder, Colorado. Association for Computational Linguistics.

\bibitem[{Karpinska et~al.(2021)Karpinska, Akoury, and Iyyer}]{karpinska2021perils}
Marzena Karpinska, Nader Akoury, and Mohit Iyyer. 2021.
\newblock \href {https://doi.org/10.18653/v1/2021.emnlp-main.97} {The perils of using {M}echanical {T}urk to evaluate open-ended text generation}.
\newblock In \emph{Proceedings of the 2021 Conference on Empirical Methods in Natural Language Processing}, pages 1265--1285, Online and Punta Cana, Dominican Republic. Association for Computational Linguistics.

\bibitem[{Kee et~al.(2018)Kee, del Castillo, and Runger}]{kee2018}
Seho Kee, Enrique del Castillo, and George Runger. 2018.
\newblock \href {https://doi.org/10.1016/j.ins.2018.05.014} {Query-by-committee improvement with diversity and density in batch active learning}.
\newblock \emph{Information Sciences}, 454–455:401--418.

\bibitem[{Li et~al.(2011)Li, Gavves, Snoek, Worring, and Smeulders}]{LiICM2011}
X.~Li, E.~Gavves, C.~G.~M. Snoek, M.~Worring, and A.~W.~M. Smeulders. 2011.
\newblock \href {https://ivi.fnwi.uva.nl/isis/publications/2011/LiICM2011} {Personalizing automated image annotation using cross-entropy}.
\newblock In \emph{ACM International Conference on Multimedia}, pages 233--242.

\bibitem[{Nowak and R\"{u}ger(2010)}]{nowak2010reliable}
Stefanie Nowak and Stefan R\"{u}ger. 2010.
\newblock \href {https://doi.org/10.1145/1743384.1743478} {How reliable are annotations via crowdsourcing: a study about inter-annotator agreement for multi-label image annotation}.
\newblock In \emph{Proceedings of the International Conference on Multimedia Information Retrieval}, MIR '10, page 557–566, New York, NY, USA. Association for Computing Machinery.

\bibitem[{Papineni et~al.(2002)Papineni, Roukos, Ward, and Zhu}]{papineni2002bleu}
Kishore Papineni, Salim Roukos, Todd Ward, and Wei-Jing Zhu. 2002.
\newblock \href {https://doi.org/10.3115/1073083.1073135} {{B}leu: a method for automatic evaluation of machine translation}.
\newblock In \emph{Proceedings of the 40th Annual Meeting of the Association for Computational Linguistics}, pages 311--318, Philadelphia, Pennsylvania, USA. Association for Computational Linguistics.

\bibitem[{Ren et~al.(2021)Ren, Xiao, Chang, Huang, Li, Gupta, Chen, and Wang}]{ren2021survey}
Pengzhen Ren, Yun Xiao, Xiaojun Chang, Po-Yao Huang, Zhihui Li, Brij~B. Gupta, Xiaojiang Chen, and Xin Wang. 2021.
\newblock \href {https://doi.org/10.1145/3472291} {{A Survey of Deep Active Learning}}.
\newblock \emph{ACM Computing Surveys}, 54(9).

\bibitem[{Saravia et~al.(2018)Saravia, Liu, Huang, Wu, and Chen}]{emotion_ref}
Elvis Saravia, Hsien-Chi~Toby Liu, Yen-Hao Huang, Junlin Wu, and Yi-Shin Chen. 2018.
\newblock \href {https://doi.org/10.18653/v1/D18-1404} {{CARER}: Contextualized affect representations for emotion recognition}.
\newblock In \emph{Proceedings of the 2018 Conference on Empirical Methods in Natural Language Processing}, pages 3687--3697, Brussels, Belgium. Association for Computational Linguistics.

\bibitem[{Sener and Savarese(2018)}]{sener2018active}
Ozan Sener and Silvio Savarese. 2018.
\newblock \href {https://openreview.net/forum?id=H1aIuk-RW} {Active learning for convolutional neural networks: A core-set approach}.
\newblock In \emph{International Conference on Learning Representations}.

\bibitem[{Socher et~al.(2013)Socher, Perelygin, Wu, Chuang, Manning, Ng, and Potts}]{sst5_ref}
Richard Socher, Alex Perelygin, Jean Wu, Jason Chuang, Christopher~D. Manning, Andrew Ng, and Christopher Potts. 2013.
\newblock \href {https://www.aclweb.org/anthology/D13-1170} {Recursive deep models for semantic compositionality over a sentiment treebank}.
\newblock In \emph{Proceedings of the 2013 Conference on Empirical Methods in Natural Language Processing}, pages 1631--1642, Seattle, Washington, USA. Association for Computational Linguistics.

\bibitem[{Touvron et~al.(2023)Touvron, Martin, Stone, Albert, Almahairi, Babaei, Bashlykov, Batra, Bhargava, Bhosale et~al.}]{touvron2023llama}
Hugo Touvron, Louis Martin, Kevin Stone, Peter Albert, Amjad Almahairi, Yasmine Babaei, Nikolay Bashlykov, Soumya Batra, Prajjwal Bhargava, Shruti Bhosale, et~al. 2023.
\newblock \href {https://arxiv.org/abs/2307.09288} {Llama 2: Open foundation and fine-tuned chat models}.
\newblock \emph{arXiv preprint arXiv:2307.09288}.

\bibitem[{Tunstall et~al.(2022)Tunstall, Reimers, Jo, Bates, Korat, Wasserblat, and Pereg}]{setfit}
Lewis Tunstall, Nils Reimers, Unso Eun~Seo Jo, Luke Bates, Daniel Korat, Moshe Wasserblat, and Oren Pereg. 2022.
\newblock \href {https://arxiv.org/abs/2209.11055} {Efficient few-shot learning without prompts}.
\newblock \emph{arXiv preprint arXiv:2209.11055}.

\bibitem[{Wolf et~al.(2020)Wolf, Debut, Sanh, Chaumond, Delangue, Moi, Cistac, Rault, Louf, Funtowicz, Davison, Shleifer, von Platen, Ma, Jernite, Plu, Xu, Le~Scao, Gugger, Drame, Lhoest, and Rush}]{wolf2020transformers}
Thomas Wolf, Lysandre Debut, Victor Sanh, Julien Chaumond, Clement Delangue, Anthony Moi, Pierric Cistac, Tim Rault, Remi Louf, Morgan Funtowicz, Joe Davison, Sam Shleifer, Patrick von Platen, Clara Ma, Yacine Jernite, Julien Plu, Canwen Xu, Teven Le~Scao, Sylvain Gugger, Mariama Drame, Quentin Lhoest, and Alexander Rush. 2020.
\newblock \href {https://doi.org/10.18653/v1/2020.emnlp-demos.6} {Transformers: State-of-the-art natural language processing}.
\newblock In \emph{Proceedings of the 2020 Conference on Empirical Methods in Natural Language Processing: System Demonstrations}, pages 38--45, Online. Association for Computational Linguistics.

\bibitem[{Yang et~al.(2023)Yang, Jin, Tang, Han, Feng, Jiang, Yin, and Hu}]{yang2023harnessing}
Jingfeng Yang, Hongye Jin, Ruixiang Tang, Xiaotian Han, Qizhang Feng, Haoming Jiang, Bing Yin, and Xia Hu. 2023.
\newblock \href {https://arxiv.org/abs/2304.13712} {Harnessing the power of llms in practice: A survey on chatgpt and beyond}.
\newblock \emph{arXiv preprint arXiv:2304.13712}.

\bibitem[{Zhang et~al.(2021)Zhang, Zhou, Jiang, Yang, Han, and Li}]{zhang2021stochastic}
Dequan Zhang, Pengfei Zhou, Chen Jiang, Meide Yang, Xu~Han, and Qing Li. 2021.
\newblock \href {https://doi.org/https://doi.org/10.1016/j.cma.2021.113990} {A stochastic process discretization method combing active learning kriging model for efficient time-variant reliability analysis}.
\newblock \emph{Computer Methods in Applied Mechanics and Engineering}, 384:113990.

\bibitem[{Zhang et~al.(2023{\natexlab{a}})Zhang, Mille, Hou, Deutsch, Clark, Liu, Mahamood, Gehrmann, Clinciu, Chandu, and Sedoc}]{zhang2023needle}
Lining Zhang, Simon Mille, Yufang Hou, Daniel Deutsch, Elizabeth Clark, Yixin Liu, Saad Mahamood, Sebastian Gehrmann, Miruna Clinciu, Khyathi~Raghavi Chandu, and Jo{\~a}o Sedoc. 2023{\natexlab{a}}.
\newblock \href {https://doi.org/10.18653/v1/2023.acl-long.835} {A needle in a haystack: An analysis of high-agreement workers on {MT}urk for summarization}.
\newblock In \emph{Proceedings of the 61st Annual Meeting of the Association for Computational Linguistics (Volume 1: Long Papers)}, pages 14944--14982, Toronto, Canada. Association for Computational Linguistics.

\bibitem[{Zhang et~al.(2015)Zhang, Zhao, and LeCun}]{agnews_ref}
Xiang Zhang, Junbo Zhao, and Yann LeCun. 2015.
\newblock \href {https://proceedings.neurips.cc/paper_files/paper/2015/file/250cf8b51c773f3f8dc8b4be867a9a02-Paper.pdf} {Character-level convolutional networks for text classification}.
\newblock In \emph{Advances in Neural Information Processing Systems}, volume~28. Curran Associates, Inc.

\bibitem[{Zhang et~al.(2023{\natexlab{b}})Zhang, Wang, Cheng, Kurohashi et~al.}]{zhang2023reformulating}
Yating Zhang, Yexiang Wang, Fei Cheng, Sadao Kurohashi, et~al. 2023{\natexlab{b}}.
\newblock \href {https://arxiv.org/abs/2310.03328} {Reformulating domain adaptation of large language models as adapt-retrieve-revise}.
\newblock \emph{arXiv preprint arXiv:2310.03328}.

\bibitem[{Zhu et~al.(2010)Zhu, Wang, Tsou, and Ma}]{zhu2010}
Jingbo Zhu, Huizhen Wang, Benjamin~K. Tsou, and Matthew Ma. 2010.
\newblock \href {https://doi.org/10.1109/TASL.2009.2033421} {Active learning with sampling by uncertainty and density for data annotations}.
\newblock \emph{IEEE Transactions on Audio, Speech, and Language Processing}, 18(6):1323--1331.

\end{thebibliography}

\end{document}